\documentclass{article} 
\usepackage{iclr2021_conference,times}

\usepackage{amsmath,amsfonts,bm}

\def\eqref#1{equation~\ref{#1}}

\def\1{\bm{1}}

\DeclareMathAlphabet{\mathsfit}{\encodingdefault}{\sfdefault}{m}{sl}
\SetMathAlphabet{\mathsfit}{bold}{\encodingdefault}{\sfdefault}{bx}{n}

\usepackage{epsfig}
\usepackage{graphicx}
\usepackage{url}            
\usepackage{booktabs}       
\usepackage{amsfonts}       
\usepackage{pifont}
\usepackage{xcolor}         
\usepackage{multirow}
\usepackage{overpic}
\usepackage{listings}
\usepackage{algorithm} 
\usepackage{hyperref}
\hypersetup{pagebackref=false,colorlinks=true,citecolor=gray,bookmarks=false,urlcolor=black}

\DeclareMathOperator{\permutemlp}{Permute-MLP}
\DeclareMathOperator{\LN}{LN}
\DeclareMathOperator{\MLP}{Channel-MLP}

\title{Vision Permutator: A Permutable MLP-Like \\ Architecture for Visual Recognition}

\author{Qibin Hou, Zihang Jiang, Li Yuan \\
Department of Electrical and Computer Engineering\\
National University of Singapore\\
Singapore \\
\texttt{\{andrewhoux,jzh0103,ylustcnus\}@gmail.com} \\
\And
Ming-Ming Cheng \\
School of Computer Science \\
Nankai University \\
Tianjin, China \\
\texttt{cmm@nankai.edu.cn} \\
\AND
Shuicheng Yan \\
Sea AI Labs \\
Singapore \\
\texttt{yansc@sea.com}
\And
Jiashi Feng \\
ECE, NUS \& Sea AI Labs \\
Singapore \\
\texttt{elefjia@nus.edu.sg} \\
}

\iclrfinalcopy 

\newcommand{\highlight}[1]{\textcolor{black}{\textbf{#1}}}
\newcommand{\myPara}[1]{\vspace{.05in}\noindent\textbf{#1:}}

\newcommand{\nameofmethod}{Vision Permutator}
\newcommand{\nameofblock}{Permutator}
\newcommand{\nameoflayer}{Permute-MLP}

\def\ie{\emph{i.e}.}
\def\eg{\emph{e.g}.}

\begin{document}

\maketitle

\begin{abstract}
In this paper, we present \nameofmethod{}, a conceptually simple and data efficient
MLP-like architecture for visual recognition.
By realizing the importance
of the positional information carried by   2D feature representations, 
unlike recent MLP-like models that encode the spatial information
along the flattened spatial dimensions, 
\nameofmethod{} 
%
separately encodes the feature representations 
along the height and width dimensions with linear projections.
This allows   \nameofmethod{} to capture long-range dependencies
along one spatial direction and meanwhile preserve precise positional information
along the other   direction. 
The resulting position-sensitive outputs are then  
aggregated in a mutually complementing manner to form expressive representations of the objects of interest.
We show that our \nameofmethod{}s are formidable competitors to convolutional neural
networks (CNNs) and vision transformers.
Without the dependence on spatial convolutions or attention mechanisms,
\nameofmethod{} achieves 81.5\% top-1 accuracy on ImageNet without
extra large-scale training data (\eg, ImageNet-22k) using only 25M learnable parameters, 
which is much better
than most CNNs and vision transformers under the same model size constraint.
When scaling up to 88M, it attains 83.2\% top-1 accuracy.
We hope this work could encourage research on rethinking the way of encoding spatial
information and facilitate the development of MLP-like models.
Code is available at \url{https://github.com/Andrew-Qibin/VisionPermutator}.
\end{abstract}

\section{Introduction}
\label{sec:introduction}

Recent studies~\citep{tolstikhin2021mlp,touvron2021resmlp} have shown that 
pure multi-layer perceptron based networks perform well in ImageNet classification~\citep{deng2009imagenet}.
Compared to convolutional neural networks (CNNs) and vision transformers that employ
spatial convolutions or self-attention layers to encode spatial information,
MLP-like networks (a.k.a., MLPs) make use of pure fully-connected layers (or called $1\times 1$ convolutions)
and hence are more efficient in both training and inference \citep{tolstikhin2021mlp}.
However, the good performance of MLPs in image classification largely benefits from training on large-scale
datasets (e.g., ImageNet-22K and JFT-300M).
Without the support of sufficiently large amount of training data, 
their performance still lags largely behind   CNNs \citep{tan2019efficientnet,brock2021high,zhang2020resnest} 
and vision transformers~\citep{jiang2021token,touvron2021going,liu2021swin}.

%
%
%

In this work, we are interested in exploiting the potential of MLPs with using merely the
ImageNet-1k data for training and target    data-efficient MLPs. To this end, we propose the \emph{\nameofmethod{}} architecture.
%
%
Specially, \nameofmethod{} innovates the existing MLP architectures 
by presenting a new layer structure that can
more effectively encode spatial information based on the basic matrix multiplication routine.
Unlike current MLP-like models, such as Mixer~\citep{tolstikhin2021mlp} 
and ResMLP~\citep{touvron2021resmlp},
that encode spatial information by flattening the spatial dimensions first and then conducting
linear projection along the spatial dimension (i.e., operating on tokens with shape ``tokens$\times$channels''), leading to the loss of positional information carried by 2D feature representations,
\nameofmethod{} maintains the original spatial dimensions of the input tokens and separately
encode spatial information along the height and width dimensions to preserve  positional information.

\begin{figure*}[t]
    \centering
    \small
    \includegraphics[width=0.8\linewidth]{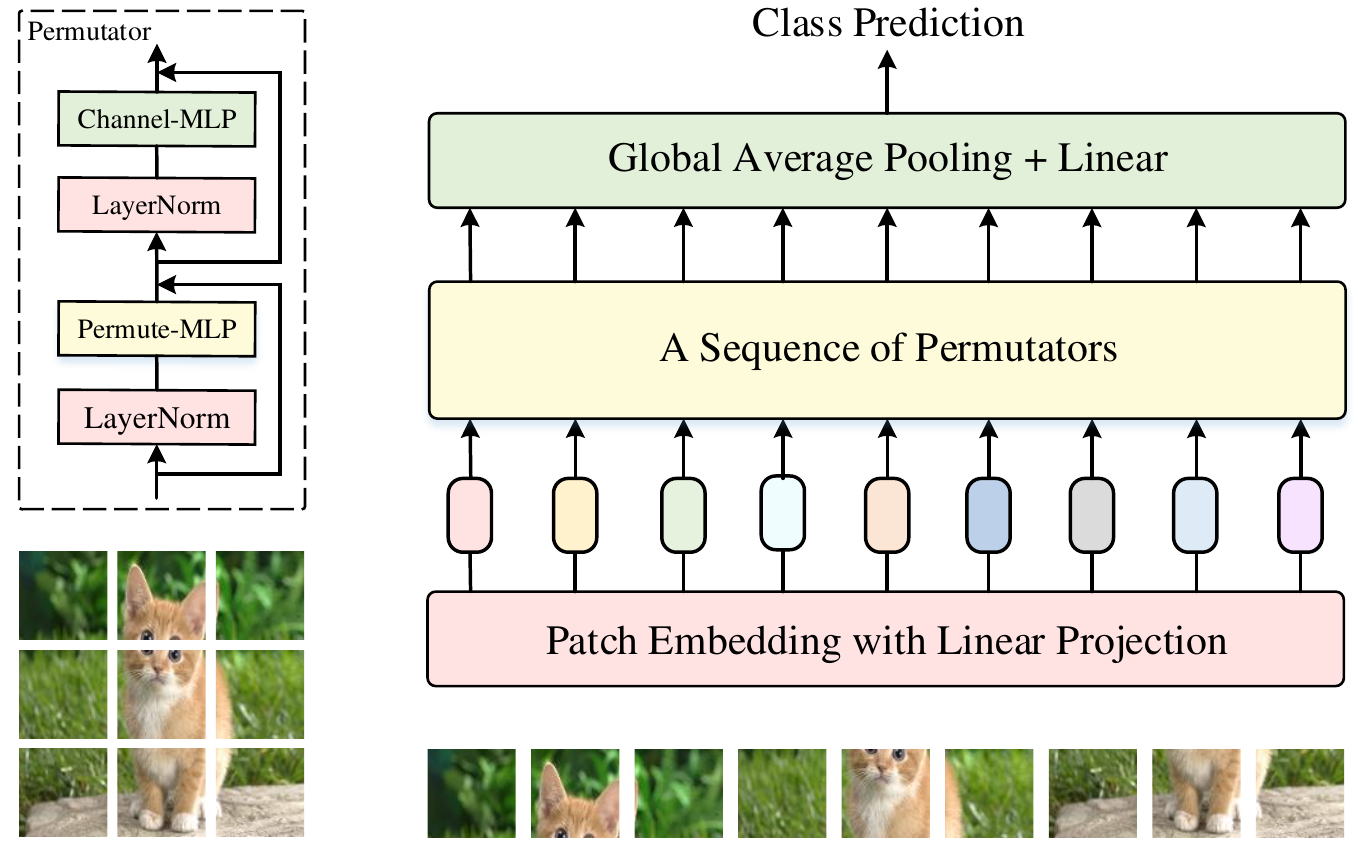}
    \caption{Basic architecture of the proposed \nameofmethod{}. The evenly divided image patches
    are tokenized with linear projection first and then fed into a sequence of \nameofblock{}s
    for feature encoding. A global average pooling layer followed by a fully-connected layer
    is finally used to predict the class.}
    \label{fig:arch}
\end{figure*}

To be specific, our \nameofmethod{} begins with a similar tokenization operation to vision transformers,
  which uniformly splits  the input image   into small   patches and then maps them to
token embeddings with linear projections,  as depicted in Figure~\ref{fig:arch}.
The resulting token embeddings with shape ``height$\times$width$\times$channels'' 
are then fed into a sequence of \nameofblock{} blocks, each of which consists
of a \nameoflayer{} for spatial information encoding and a Channel-MLP
for channel information mixing.
%
The \nameoflayer{} layer, as depicted in Figure~\ref{fig:permute_mlp}, consists of three independent
branches, each of which encodes  features along  a specific dimension, \ie, the height,     width or
  channel dimension.
%
%
Compared to existing MLP-like models that mix the two spatial dimensions into one, our \nameofmethod{}
separately processes the token representations along these dimensions, resulting in
tokens with direction-specific information, which has been demonstrated essential for
visual recognition \citep{hou2021coordinate,wang2020axial}.

Experiments show that our \nameofmethod{} can largely improve the classification performance of 
existing MLP-like models.
Taking the small-sized \nameofmethod{} as an example, it attains 81.5\% top-1
accuracy on ImageNet without any extra training data.
Scaling up the model to 55M and 88M, we can further achieve 82.7\% and 83.2\%
accuracy, respectively.

\section{Related Work} \label{sec:related_work}

Modern deep neural networks for image classification can be mainly categorized into three
different classes: convolutional neural networks (CNNs), vision transformers (ViTs),
and multi-layer perceptron based models (MLPs).
In the following, we will briefly describe the development trend of each type of networks and
state the differences of the proposed \nameofmethod{} from previous work.

CNNs, as the de-facto standard networks in computer vision for years, have been deeply studied.
Early CNN models, such as AlexNet~\citep{krizhevsky2012imagenet} and VGGNet~\citep{simonyan2014very},
mostly adopt structures with a stack of spatial convolutions (with kernel size $\geq 3$) and pooling
operations.
Later, ResNets and their variants~\citep{he2016deep,xie2017aggregated,zagoruyko2016wide} introduce
skip connection and building blocks with bottleneck structure into CNNs, 
enabling training very deep networks possible.
Inceptions~\citep{szegedy2015going,szegedy2016rethinking} renovate the design of traditional building
block structure and utilize multiple parallel paths of sets of specialized filters.
Attention mechanisms \citep{hu2018squeeze,hu2019local,wang2018non,bello2019attention,liu2020improving,chen20182}
break through the limitations of convolutions in capturing local features and further promote
the development of CNNs.
Our work can also be regarded as a special CNN.
Different from previous CNNs that globally aggregate the locally captured features
with spatial convolutions, our \nameofmethod{} is composed of pure $1\times1$ convolutions
but can encode global information.

Our work is also related to vision transformers \citep{dosovitskiy2020image}.
Unlike CNNs that exploit local convolutions to encode spatial information,
vision transformers takes advantage of the self-attention mechanism to capture global information
and have been the prevailing research direction in image classification recently.
Since then, a great number of transformer-based classification models appear, aiming at 
advancing the original vision transformer by either introducing locality \citep{zhou2021refiner,vaswani2021scaling,wu2021cvt,liu2021swin,han2021transformer,yuan2021tokens}, 
or scaling the depth \citep{zhou2021deepvit,touvron2021going}, or tailoring powerful optimization strategies~\citep{jiang2021token}.
%
%
Different from the aforementioned methods, our \nameofmethod{} eliminates
the dependence on self-attention and hence is more efficient.

Very recently, there are also some work \citep{tolstikhin2021mlp,touvron2021resmlp,liu2021pay,guo2021beyond} 
targeting at developing pure MLP-like models for ImageNet classification.
To encode rich spatial information with MLPs, these methods flatten the spatial dimensions and 
treat the three-dimensional (height, width, and channel) token representations as a two-dimensional
input table.
Differently, our \nameofmethod{} operates on three-dimensional feature representations and
encodes spatial information separately along the height and width dimensions.
We will show the advantages of the proposed \nameofmethod{} over existing MLP-like models
in our experiment section.

\section{\nameofmethod{}} \label{sec:method}

The basic architecture of the proposed \nameofmethod{} can be found in Figure~\ref{fig:arch}.
Our network takes an image of size $224\times224$ as input and uniformly
splits it into a sequence of image patches ($14\times14$ or $7\times7$).
All the patches are then mapped into linear embeddings (or called tokens) using a
shared linear layer as \citep{tolstikhin2021mlp}.
We next feed all the tokens into a sequence of Permutators to encode both
spatial and channel information.
The resulting tokens are finally averaged along the spatial dimensions, followed by
a fully-connected layer for class prediction.
In the following, we will detail the proposed Permutator block and the
network settings.

\begin{figure*}[t]
    \centering
    \small
    \includegraphics[width=\linewidth]{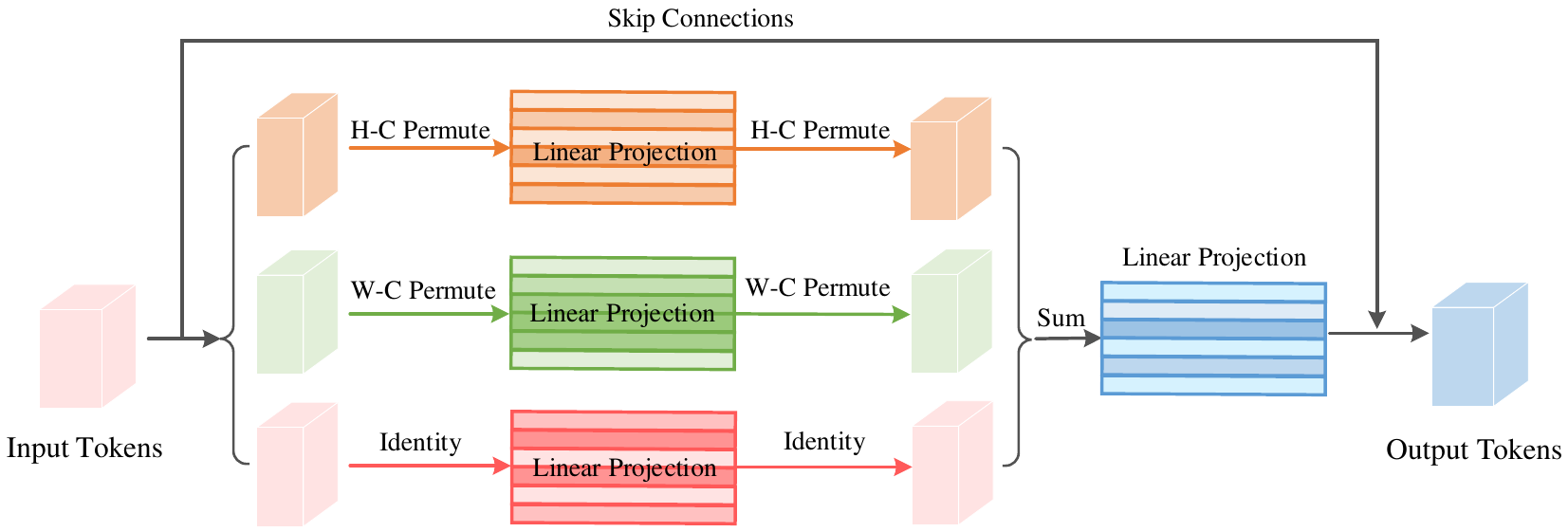}
    \caption{Basic structure of the proposed \nameoflayer{} layer. The proposed \nameoflayer{} layer contains
    three branches that are responsible for encoding features along the height, width, and channel
    dimensions, respectively. The outputs from the three branches are then combined using element-wise addition, followed by a fully-connected layer for feature fusion.}
    \label{fig:permute_mlp}
\end{figure*}

\subsection{Permutator}

A diagrammatic illustration of the proposed \nameofblock{} block can be found at top-left corner
of Figure~\ref{fig:arch}.
As can be seen, regardless of the LayerNorms and the skip connections, 
our Permutator consists of two components: Permute-MLP and Channel-MLP,
which are responsible for encoding spatial information and channel information, respectively.
The Channel-MLP module shares a similar structure to the feed forward layer 
in Transformers \citep{vaswani2017attention} that comprises two fully-connected layers
with a GELU activation in the middle.
For spatial information encoding, unlike the recent Mixer \citep{tolstikhin2021mlp} that
conducts linear projection along the spatial dimension with respect to all the tokens,
we propose to separately processing the tokens along the height and width dimensions.
Mathematically, given an input $C$-dim tokens $\mathbf{X} \in \mathbb{R}^{H \times W \times C}$, 
the formulation of \nameofblock{} can be written as follows:
\begin{align}
    \mathbf{Y} &= \permutemlp(\LN(\mathbf{X})) + \mathbf{X}, \\
    \mathbf{Z} &= \MLP(\LN(\mathbf{Y})) + \mathbf{Y},
\end{align}
where, $\LN$ refers to LayerNorm.
The output $\mathbf{Z}$ will serve as the input to the next \nameofblock{} block until the last one.

\begin{algorithm}[t]
\caption{{Code for Permute-MLP (PyTorch-like)}}
\label{alg:permute_mlp}
\definecolor{codeblue}{rgb}{0.25,0.5,0.25}
\lstset{
	backgroundcolor=\color{white},
	basicstyle=\fontsize{7.2pt}{7.2pt}\ttfamily\selectfont,
	columns=fullflexible,
	breaklines=true,
	captionpos=b,
	commentstyle=\fontsize{7.2pt}{7.2pt}\color{codeblue},
	keywordstyle=\fontsize{7.2pt}{7.2pt},
}
\begin{lstlisting}[language=python]
# H: height, W: width, C: channel, S: number of segments
# x: input tensor of shape (H, W, C)

################### initialization ####################################################
proj_h = nn.Linear(C, C)    # Encoding spatial information along the height dimension
proj_w = nn.Linear(C, C)    # Encoding spatial information along the width dimension
proj_c = nn.Linear(C, C)    # Encoding channel information
proj   = nn.Linear(C, C)    # For information fusion

#################### code in forward ##################################################
def permute_mlp(x):
    N = C // S 
    x_h = x.reshape(H, W, N, S).permute(2, 1, 0, 3).reshape(N, W, H*S)
    x_h = self.proj_h(x_h).reshape(N, W, H, S).permute(2, 1, 0, 3).reshape(H, W, C)

    x_w = x.reshape(H, W, N, S).permute(0, 2, 1, 3).reshape(H, N, W*S)
    x_w = self.proj_w(x_w).reshape(H, N, W, S).permute(0, 2, 1, 3).reshape(H, W, C)

    x_c = self.proj_c(x)

    x = x_h + x_w + x_c
    x = self.proj(x)
    return x
\end{lstlisting}
\end{algorithm}

\myPara{\nameoflayer{}} 
The visual illustration of the proposed \nameoflayer{} can be found in Figure~\ref{fig:permute_mlp}.
Unlike vision transformers \citep{dosovitskiy2020image,jiang2021token,touvron2020training}
and Mixer \citep{tolstikhin2021mlp} that receive an input of two dimensions (``tokens$\times$channels,''
\ie, $HW \times C$), \nameoflayer{} accepts 3-dimensional token representations.
As shown in Figure~\ref{fig:permute_mlp}, our \nameoflayer{} consists of three branches, each of which
is in charge of encoding information along the either height, or width, or channel dimension.
The channel information encoding is simple as we only need a fully-connected layer with weights
$\mathbf{W}_C \in \mathbb{R}^{C \times C}$ to perform a linear projection with respect to the input
$\mathbf{X}$, yielding $\mathbf{X}_C$.
In the following, we will describe how to encode spatial information by introducing
a head-wise permutation operation between dimensions.

%
Suppose the hidden dimension $C$ is 384 and the input image is with resolution $224\times224$.
To encode the spatial information along the height dimension, we first conduct a
height-channel permutation operation.
Given the input $\mathbf{X} \in \mathbb{R}^{H \times W \times C}$, we first split it into $S$ segments
along the channel dimension, yielding $[\mathbf{X}_{H_1}, \mathbf{X}_{H_2}, \cdots, \mathbf{X}_{H_S}]$,
satisfying $C = N * S$\footnote{In our case, $N$ is identical to $H$ or $W$.}.
In case where the patch size is set to $14\times14$, the value of $N$ is identical to 16 and
$\mathbf{X}_{H_i} \in \mathbb{R}^{H \times W \times N}, (i \in \{1, \cdots, S\})$.
We then perform a height-channel permutation operation\footnote{Transpose the first (\textbf{H}eight) dimension and the third (\textbf{C}hannel) dimension: $(H, W, C) \rightarrow (C, W, H)$.} with respect to
each segment $\mathbf{X}_{H_i}$, yielding $[\mathbf{X}_{H_1}^\top, \mathbf{X}_{H_2}^\top, \cdots, \mathbf{X}_{H_S}^\top]$, which are then concatenated along the channel dimension as the output
the permutation operation.
Next, a fully-connected layer with weight $\mathbf{W}_H \in \mathbb{R}^{C \times C}$ is connected 
to mix the height information.
To recover the original dimensional information to $\mathbf{X}$, we only need to perform
the height-channel permutation operation once again, outputting $\mathbf{X}_H$.
Similarly, in the second branch, we conduct the same operations as above to permute the
width dimension and the channel dimension for $\mathbf{X}$ and yield $\mathbf{X}_W$.
Finally, we feed the summation of  all the token representations from the three branches
into a new fully-connected layer to attain the output of the \nameoflayer{} layer, which
can be formulated as follows:
\begin{equation} \label{eqn:fusion}
    \hat{\mathbf{X}} = \text{FC}(\mathbf{X}_H + \mathbf{X}_W + \mathbf{X}_C),
\end{equation}
where $\text{FC}(\cdot)$ denotes a fully-connected layer with weight $\mathbf{W}_P \in \mathbb{R}^{C \times C}$.
A PyTorch-like pseudo code can be found in Alg.~\ref{alg:permute_mlp}.

\myPara{Weighted \nameoflayer{}} 
In Eqn.~\ref{eqn:fusion}, we simply fuse the outputs
from all three branches with element-wise addition.
Here, we further improve the above \nameoflayer{} by recalibrating the importance
of different branches and present Weighted \nameoflayer{}.
This can be easily implemented by exploiting the split attention \citep{zhang2020resnest}.
What is different is that the split attention is applied to $\mathbf{X}_H$, $\mathbf{X}_W$,
and $\mathbf{X}_C$ instead of a group of tensors generated by a grouped convolution.
In the following, we use the weighted \nameoflayer{} in \nameofblock{} by default.

\subsection{Various Configurations of \nameofmethod{}}

We summarize various configurations of the proposed \nameofmethod{} in Table~\ref{tab:config}.
We present three different versions of \nameofmethod{} (ViP), denoted as `ViP-Small', `ViP-Medium', and
`ViP-Large' respectively, according to their model size.
Notation `ViP-Small/14' denotes the small-sized model with patch size $14\times14$
in the starting patch embedding module.
In `ViP-Small/16' and `ViP-Small/14', there is only one patch embedding module, which
is then folloed by a sequence of \nameofblock{}s.
The total number of \nameofblock{}s for them are 16.

Our `ViP-Small/7,' `ViP-Medium/7,' and `ViP-Large/7' have two stages, each of which starts with a
patch embedding module.
For these models, we add a few \nameofblock{}s targeting at encoding fine-level token representations
which we found beneficial to the model performance.
In our experiment section, we will show the advantage of encoding fine-level token representations.


\begin{table*}[t]
  \centering
  \small
  \setlength\tabcolsep{2.8mm}
  \renewcommand\arraystretch{1}
  \caption{Configurations of different \nameofmethod{} models. We present three different models (Small, Medium, and Large) according to the different model sizes. Notation ``Small/16'' means the model with patch size
  $16\times16$ in the starting patch embedding module.}
  \label{tab:config}
  \begin{tabular}{lcccccccc} \toprule
    Specification & ViP-Small/16 & ViP-Small/14 & ViP-Small/7 & ViP-Medium/7 & ViP-Large/7 \\ \midrule
    
    Patch size & $16\times16$ & $14\times14$ & $7\times7$ & $7\times7$ & $7\times7$ \\
    Hidden size & - & - & 192 & 256 & 256 \\
    \#Tokens & $14\times14$ & $16\times16$ & $32\times32$ & $32\times32$ & $32\times32$ \\
    \#\nameofblock{}s & - & - & 4 & 7 & 9  \\   \midrule
    Patch size & - & - & $2\times2$ & $2\times2$ & $2\times2$ \\
    Hidden size & 336 & 384 & 384 & 512 & 512 \\
    \#Tokens & $14\times14$ & $16\times16$ & $16\times16$ & $16\times16$ & $16\times16$ \\
    \#\nameofblock{}s & 18 & 18 & 14 & 17 & 27 \\
    \midrule
    Number of layers & 18 & 18 & 18 & 24 & 36 \\
    MLP Ratio & 3 & 3 & 3 & 3 & 3 \\
    Stoch. Dep. & 0.1 & 0.1 & 0.1 & 0.2 & 0.3 \\
    Parameters (M) & 23M & 30M & 25M & 55M & 88M \\
    \bottomrule
  \end{tabular}
\end{table*}

\section{Experiments}

We report of the results of our proposed \nameofmethod{} on the widely-used 
ImageNet-1k \citep{deng2009imagenet} dataset.
The code is implemented based on \texttt{PyTorch}~\citep{paszke2019pytorch} and 
the \texttt{timm}~\citep{rw2019timm} toolbox.
Note that in training, we do not use any extra training data.

\subsection{Experiment Setup}

We adopt the AdamW optimizer~\citep{loshchilov2017decoupled} with a linear learning rate scaling strategy 
$lr = 10^{-3} \times \frac{batch\_size}{1024}$ and $5\times 10^{-2}$ weight decay rate to optimize
all the models as suggested by previous work \citep{touvron2020training,jiang2021token}. 
The batch size is set to 2048 which we found works better than 1024 in our \nameofmethod{}.
%
%
%
%
Stochastic Depth~\citep{huang2016deep} is used.
Detailed drop rates can be found in Table~\ref{tab:config}.
We train our models on the ImageNet dataset for 300 epochs.
%
For data augmentation methods, we use CutOut~\citep{zhong2020random}, RandAug~\citep{cubuk2020randaugment},
MixUp~\citep{zhang2017mixup}, and CutMix~\citep{yun2019cutmix}.
%
%
Note that we do not use positional encoding in our \nameofmethod{} as we found
it hurts the performance.
Training small-sized \nameofmethod{} models requires a machine node with 8 NVIDIA V100 GPUs (32G memory).
Two nodes are needed for medium-sized and large-sized \nameofmethod{} models.
%
%
%

\subsection{Main Results on ImageNet}

In this subsection, we compare our proposed \nameofmethod{} with previous CNN-based,
Transformer-based, and MLP-like models.
We first compare our proposed \nameofmethod{} with recent MLP-like models in Table~\ref{tab:comp_mlps}.
The `Train size' and `Test Size' refer to the training resolution and test resolution, respectively.
Our ViP-Small/7 model with 25M parameters achieves top-1 accuracy of 81.5\%.
This result is already better than most of the existing MLP-like models and
comparable to the best one gMLP-B~\citep{liu2021pay} with 73M parameters.
Scaling up the model to 55M allows our ViP-Medium/7 to attain 82.7\% accuracy, which
is better than all other MLP-like models as shown in Table~\ref{tab:comp_mlps}.
Further increasing the model size to 88M leads to a better result 83.2\%.

\begin{table*}[t]
    \centering
    \setlength\tabcolsep{1.6mm}
    \caption{Top-1 accuracy comparison with recent MLP-like models on ImageNet \citep{deng2009imagenet}.
    All models are trained without external data. 
    With the same computation and parameter constraint, our model consistently outperforms
    other methods. Following~\citep{touvron2021resmlp}, the throughput is measured on a single 
    machine with V100 GPU (32GB) with batch size set to 32. $^\dagger$ Implementation
    with our training recipe, which we found works better than the one reported in the paper.
    }
    \label{tab:comp_mlps}
    \def \mysp {\hspace{7pt}}
    {\small 
    \begin{tabular}{l|@{\ }@{\ }cc|cc|c}
    \toprule
    Networks & Parameters & Throughput & Train size & Test size  &  Top-1  Acc. (\%) \\
    \toprule
    EAMLP-14~\citep{guo2021beyond} & 30M & 711 img/s & 224 & 224 & 78.9 \\
    gMLP-S~\citep{liu2021pay} & 20M & - & 224 & 224& 79.6 \\
    ResMLP-S24~\citep{touvron2021resmlp} & 30M & 715 img/s& 224 & 224 & 79.4  \\
    ViP-Small/14 (ours)  &  30M & 789 img/s & 224 & 224 & 80.5 \\
    ViP-Small/7 (ours) &  25M & 719 img/s & 224 & 224 & \highlight{81.5} \\ \toprule
    EAMLP-19~\citep{guo2021beyond} & 55M & 464 img/s & 224 & 224 & 79.4 \\
    Mixer-B/16~\citep{tolstikhin2021mlp}$^\dagger$ & 59M & - & 224 & 224 & 78.5  \\
    ViP-Medium/7 (ours) & 55M & 418 img/s & 224 & 224 & \highlight{82.7} \\ \toprule
    gMLP-B~\citep{liu2021pay} & 73M & - & 224 & 224 & 81.6 \\
    ResMLP-B24~\citep{touvron2021resmlp} & 116M & 231 img/s & 224 & 224 & 81.0  \\
    ViP-Large/7 (ours) &  88M & 298 img/s & 224 & 224 & \highlight{83.2} \\
    \bottomrule
    \end{tabular}}
\end{table*}

\begin{table*}[htp!]
    \centering
    \setlength\tabcolsep{1.8mm}
    \caption{Top-1 accuracy comparison with classic CNNs and Vision Transformers on ImageNet \citep{deng2009imagenet}.
    All models are trained without external data. 
    With the same computation and parameter constraint, our models are competitive to
    some powerful CNN-based and transformer-based counterparts.
    }
    \label{tab:comp_cnns}
    \def \mysp {\hspace{7pt}}
    {\small 
    \begin{tabular}{l|@{\ }@{\ }c|cc|c}
    \toprule
    Network & Parameters  & Train size & Test size  &  Top-1  Acc. (\%) \\
    \toprule

    ResNet-50d~\citep{he2016deep,he2019bag} & 25.6M & 224 & 224 & 79.5  \\
    SE-ResNeXt-50~\citep{xie2017aggregated,hu2018squeeze} & 27.6M & 224 & 224 & 79.9  \\
    RegNet-6.4GF~\citep{radosavovic2020designing} & 30.6M & 224 & 224 & 79.9 \\
    ResNeSt-50~\citep{zhang2020resnest}  & 27.5M & 224 & 224 & 81.1  \\
    DeiT-S~\citep{touvron2020training} & 22M & $224$ & $224$ & 79.9  \\
    T2T-ViT-14~\citep{yuan2021tokens}  & 22M & $224$ & $224$ & 81.5  \\
    Swin-T~\citep{liu2021swin} & 29M & 224 & 224 & 81.3 \\
    ViP-Small/7 &  25M & 224 & 224 & 81.5 \\ \toprule
    ResNet-101d~\citep{he2016deep,he2019bag} & 44.6M & 224 & 224 & 80.4  \\
    SE-ResNeXt-101~\citep{xie2017aggregated,hu2018squeeze} & 49.0M & 224 & 224 & 80.9  \\
    RegNet-12GF~\citep{he2016deep,he2019bag} & 51.8M & 224 & 224 & 80.3 \\
    ResNeSt-101~\citep{zhang2020resnest}  & 48.3M & 256 & 256 & 82.9  \\
    DeepViT~\citep{zhou2021deepvit} & 55M & 224 & 224 & 83.1 \\
    ViP-Medium/7 & 55M & 224 & 224 & 82.7 \\
    \toprule
    RegNet-16GF~\citep{radosavovic2020designing} & 83.6M & 224 & 224 & 80.4 \\
    DeiT-B~\citep{touvron2020training} & 86M & $224$ & $224$ & 81.8  \\
    T2T-ViT-24~\citep{yuan2021tokens}  & 64M & $224$ & $224$ & 82.3  \\
    TNT-B~\citep{han2021transformer} & 66M & 224 & 224 & 82.8 \\
    ViP-Large/7 &  88M & 224 & 224 & 83.2 \\
    \bottomrule
    \end{tabular}}
\end{table*}

We argue that the main factor leading to the improvement for our \nameofmethod{}
is the way of encoding spatial information as described in Sec.~\ref{sec:method}.
Different from concurrent popular MLP-like models listed in Table~\ref{tab:comp_mlps},
we separately encoding the token representations along the height and width dimensions,
generating position-sensitive outputs that are crucial for locating and 
identifying objects of interest~\citep{hou2021coordinate,wang2020axial}.
In addition, our \nameofmethod{} encodes not only coarse-level token representations
(with $16\times16$ tokens) but also features at fine-level (with $32\times32$ tokens).
We will detail this in next subsection.


In Table~\ref{tab:comp_cnns}, we show the comparison with classic CNN-based 
and transformer-based models.
%
Compared with classic CNNs, like ResNets~\citep{he2016deep}, SE-ResNeXt~\citep{xie2017aggregated,hu2018squeeze}, and RegNet~\citep{radosavovic2020designing}, our \nameofmethod{}
with similar model size constraint receives better results.
Taking the ViP-Small/7 model as an example, the performance is 81.5\%,
which is even better than ResNeSt-50 (81.5\% \emph{v.s.} 81.1\%).
Compared to some transformer-based models, such as DeiT~\citep{touvron2020training},
T2T-ViT~\citep{yuan2021tokens}, and Swin Transformers~\citep{liu2021swin},
our results are also better.
However, there is still a large gap between our \nameofmethod{} and recent
state-of-the-art CNN- and transformer-based models, such as NFNet~\citep{brock2021high} (86.5\%),
LV-ViT~\citep{jiang2021token} (86.4\%), and CaiT~\citep{touvron2021going} (86.5\%).
We believe there is still a large room for improving MLP-like models,
just like what happened in the research field of vision transformers.

\subsection{Ablation Analysis}

In this subsection, we conduct a series of ablation experiments on fine-level information encoding, model scaling, data augmentation, and the proposed \nameofblock{}.
We take the ViP-Small/14 model as baseline.

\begin{table}[h]
  \centering
  \small
  \setlength\tabcolsep{2mm}
  \renewcommand\arraystretch{1}
  \caption{Role of fine-level token representation encoding. `Initial Patch Size'
  denotes the patch size in the starting patch embedding module and `Fine Tokens' refers to models encoding fine-level token representations. Larger patch size means
  that the number of tokens fed into \nameofblock{}s would be lower as specified in Table~\ref{tab:config}. We can see that the model efficiency in speed does not
  change too much when changing the initial patch size.}
  \label{tab:abl_fine}
  \begin{tabular}{lcccccc} \toprule
    Models & Initial Patch Size & Fine Tokens & Layers & Parameters & Throughput & Top-1 Acc. (\%) \\ \midrule[0.5pt] 
    ViP-Small/16 & $16\times16$ & No & 18 & 23M & 803 img/s & 79.8 \\
    ViP-Small/14 & $14\times14$ & No & 18 & 30M & 789 img/s & 80.6 \\
    ViP-Small/7  & $7\times7$   & Yes & 18 & 25M   & 719 img/s & 81.5 \\
    \bottomrule
  \end{tabular}
\end{table}

\myPara{Importance of Fine-level Token Representation Encoding}
We first show that encoding finer-level token representations is important for
MLP-like models.
We demonstrate this argument in two ways: I) Adjusting the patch size in the initial
patch embedding layer and keep the backbone unchanged; II) Halving the patch size
for each patch side and introducing a few \nameofblock{}s to encode fine-level token
representations.
Table~\ref{tab:abl_fine} summaries the performance for ViP-Small/16, ViP-Small/14,
and ViP-Small/7.
Compared to ViP-Small/16, ViP-Small/14 has smaller initial patch size and
more input tokens to the \nameofblock{}s.
According to the results, ViP-Small/14 yields better performance than ViP-Small/16 (80.5\% \emph{v.s.} 79.8\%).
Despite more tokens and more parameters used in ViP-Small/14, 
the efficiency (throughput) does not change much.
This indicates that we can appropriately use smaller initial patch size to
improve the model performance.

We further reduce the initial patch size from $14\times14$ to $7\times7$.
Compared to ViP-Small/14, ViP-Small/7 adopts 4 \nameofblock{}s to 
encode fine-level token representations (with $32\times32$ tokens).
As shown in Table~\ref{tab:abl_fine}, such a slight modification can largely
boost the performance and reduce the number of learnable parameters.
The top-1 accuracy is improved from 80.5\% to 81.5\%.
This demonstrates that encoding fine-level token representations does help
in improving our model performance but
a disadvantage is that the efficiency goes down a little.
%

\begin{table}[h]
  \centering
  \small
  \setlength\tabcolsep{2.2mm}
  \renewcommand\arraystretch{1}
  \caption{Role of the model scale. We scale the models by increasing the model
  size (including number of layers, hidden dimension). `Hidden Dim.' refers to
  the hidden dimension in the second stage, which is halved in the first stage.
  Clearly, increasing the model size can consistently improve the model performance.}
  \label{tab:abl_scaling}
  \begin{tabular}{lcccccc} \toprule
    Models & Layers & Hidden Dim. & Fine Tokens & Parameters & Throughput & Top-1 Acc. (\%) \\ \midrule[0.5pt] 
    ViP-Small/7  & 18 & 384 & Yes & 25M & 719 img/s & 81.5 \\
    ViP-Medium/7 & 24 & 512 & Yes & 55M & 418 img/s & 82.7 \\
    ViP-Large/7  & 36 & 512 & Yes & 88M & 298 img/s & 83.2 \\
    \bottomrule
  \end{tabular}
\end{table}

\myPara{Role of the model scale}
Scaling up models for deep neural networks is always an effective way to improve
model performance.
Here, we show the influence of model scaling on the proposed \nameofmethod{} by increasing
the number of layers and hidden dimension.
Table~\ref{tab:abl_scaling} lists the results for three different versions of the proposed
\nameofmethod{}: ViP-Small/7, ViP-Medium/7, and ViP-Large/7.
We can see that increasing the number of layers and hidden dimension yields better
results for our \nameofmethod{}.
The ViP-Medium/7 can raise the performance of ViP-Small/7 to 82.7\% with a performance
gain of more than 1\%.
Further increasing the model size results in better performance 83.2\%.

\myPara{Effect of Data Augmentations} Data augmentation has been demonstrated
an effective and efficient way to lift the model performance 
in deep learning ~\citep{he2019bag,touvron2020training,jiang2021token}.
Four commonly-used data augmentation methods should be Random Augmentation~\citep{cubuk2020randaugment}, CutOut~\citep{zhong2020random}, MixUp~\citep{zhang2017mixup}, and CutMix~\citep{yun2019cutmix}.
Here, we show how each method influences the model performance.
The results have been shown in Table~\ref{tab:abl_aug}.
Without any data augmentation, we achieve 75.3\% top-1 accuracy for
our ViP-Small/14 model.
Using Random Augmentation improves the performance to 77.7\% (+2.4\%).
Adding CutOut lifts the result to 78.0\% (+2.7\%).
Adding MixUp yields 80.2\% top-1 accuracy (+4.9\%) and
the result is further improved to 80.6\% (+5.3\%) by using CutMix.
These experiments indicate that data augmentation is extremely important
in training \nameofmethod{} as happened in training CNNs~\citep{he2019bag}
and vision transformers~\citep{touvron2020training,jiang2021token}.

\begin{table}[h]
  \centering
  \small
  \setlength\tabcolsep{3mm}
  \renewcommand\arraystretch{1}
  \caption{Ablation on data augmentation methods. We ablate four widely used data augmentation methods in both CNN- and transformer-based models, including Random Augmentation~\citep{cubuk2020randaugment}, CutOut~\citep{zhong2020random}, MixUp~\citep{zhang2017mixup}, and CutMix~\citep{yun2019cutmix}. We can see that all 4 methods contribute to the model performance.}
  \label{tab:abl_aug}
  \begin{tabular}{lccccc} \toprule
    Data augmentation methods in training & Layers & Parameters & Top-1 Acc. (\%) \\ \midrule[0.5pt] 
    Baseline (ViP-Small/14)   & 16 & 30M & 75.3 \\
    + Random Augmentation~\citep{cubuk2020randaugment}   & 18 & 30M & 77.7 (\highlight{+2.4}) \\
    + CutOut~\citep{zhong2020random}                     & 18 & 30M & 78.0 (\highlight{+2.7})\\
    + MixUp~\citep{zhang2017mixup}                       & 18 & 30M & 80.2 (\highlight{+4.9}) \\
    + CutMix~\citep{yun2019cutmix}                       & 18 & 30M & 80.6 (\highlight{+5.3})\\
    \bottomrule
  \end{tabular}
\end{table}

\myPara{Ablation on \nameofblock{}} In this paragraph, we demonstrate the importance
of encoding spatial information along the height and width dimensions separately and
show how weighted \nameofblock{} helps in improving model performance.
In Table~\ref{tab:abl_permutator}, we summarize the results under different \nameofblock{}
settings.
Detailed description on each setting can be found in the caption.
We can see that discarding either height information encoding or width information
encoding leads to worse performance (80.2\% \emph{v.s.} 72.8\% or 72.7\%).
This demonstrates that encoding both height and width information is important.
In addition, we can also observe that replacing the vanilla \nameoflayer{}
with the Weighted \nameoflayer{} can further improve the performance
from 80.2\% to 80.6\%.

\begin{table}[h]
  \centering
  \small
  \setlength\tabcolsep{2.8mm}
  \renewcommand\arraystretch{1}
  \caption{Ablation on \nameofmethod{}. `ViP-Small/14 w/o Height' means a ViP-Small/14
  model with the height information encoding part replaced by channel encoding (the bottom
  branch in Figure~\ref{fig:permute_mlp}). A similar meaning holds for `ViP-Small/14 w/o Width.' `ViP-Small/14 w/ Permute-MLP' refers to model with the vanilla \nameoflayer{}.}
  \label{tab:abl_permutator}
  \begin{tabular}{lccccc} \toprule
    Model Specification & Layers & Parameters & Throughput & Top-1 Acc. (\%) \\ \midrule[0.5pt] 
    ViP-Small/14 w/o Height Information  & 18 & 29M &  844 img/s & 72.8 (\highlight{-7.8})\\
    ViP-Small/14 w/o Width Information & 18 & 29M & 843 img/s  & 72.7 (\highlight{-7.9})\\
    ViP-Small/14 w/ Permute-MLP & 18 & 29M & 847 img/s & 80.2 (\highlight{-0.4})\\
    ViP-Small/14 w/ Weighted Permute-MLP  & 18 & 30M & 789 img/s  & 80.6 \\
    \bottomrule
  \end{tabular}
\end{table}



\section*{Conclusions and Future Work}

In this paper, we present a novel MLP-like network architecture for visual recognition,
termed \nameofmethod{}.
We demonstrate that separately encoding the height and width information can largely
improve the model performance compared to recent MLP-like models that deem
the two spatial dimensions as one.
Our experiments also give full support of this.

Despite the large improvement over concurrent popular MLP-like models,
a clear downside of the proposed \nameofblock{} is the scaling problem
in spatial dimensions, which also exists in other MLP-like models.
As the shapes of the parameters in fully-connected layers are fixed,
it is impossible to process input images with arbitrary shapes.
This makes MLP-like models difficult to be used in down-stream tasks
with various-sized input images.

Our future work will be continuously put on the development of MLP-like models
considering the high efficacy in parallelization.
Specifically, we will continue to conquer the limitations of MLP-like models
in processing input images with arbitrary shapes and their applications in 
down-stream tasks, such as object detection and semantic segmentation.

\bibliography{egbib}
\bibliographystyle{iclr2021_conference}

\end{document}